%% file: main.tex
\documentclass[lettersize,journal]{IEEEtran}

\usepackage[noend]{algpseudocode}
\usepackage{algorithmicx,algorithm}
\usepackage{cite}
\usepackage{tikz}
\usepackage{amsmath}
\usepackage{comment}
\usepackage{color}
\usepackage{booktabs}
\usepackage{xspace}
\usepackage{amssymb}
\usepackage{amsfonts}           
\usepackage{mathrsfs} 
\usepackage{subfigure}
\usepackage{multirow}
\usepackage{booktabs}
\usepackage{graphicx}
\usepackage{enumitem}
\usepackage{setspace}
\usepackage{caption} 
\usepackage{bbding}
\newcommand{\cmark}{\ding{51}}%
\newcommand{\xmark}{\ding{55}}%
\usepackage{pifont}
\usepackage{bbm}

\usepackage{hyperref}
\hypersetup{hidelinks} 

\begin{document}

\title{Semantic-Aware Local-Global Vision Transformer}

\author{Jiatong Zhang\textsuperscript{\dag}, 
        Zengwei Yao\textsuperscript{\dag}, 
        Fanglin Chen,~\IEEEmembership{Member,~IEEE},
        Guangming Lu,~\IEEEmembership{Member,~IEEE},
        and
        Wenjie Pei$^*$
\thanks{\textsuperscript{\dag} Equal contribution.}
\thanks{$^*$ Wenjie Pei is the corresponding author.}
\thanks{Jiatong Zhang, Fanglin Chen, Guangming Lu and Wenjie Pei are with the Department of
Computer Science, Harbin Institute of Technology at Shenzhen, Shenzhen 518057, China (e-mail: zhjiatong@163.com; chenfanglin@hit.edu.cn; luguangm@hit.edu.cn; wenjiecoder@outlook.com).}%
\thanks{Zengwei Yao is with Xiaomi Corporation, Beijing, China (e-mail: yaozengwei@xiaomi.com).}
}



\maketitle

\begin{abstract}
\input{abstract}
\end{abstract}


\section{Introduction}
\input{Introduction}

\section{Related Work}
\input{Related-work}

\section{Method}
\label{sec:method}
\input{Method}

\section{Experiment}
\label{sec:Experiment}

\input{Experiment}

\section{Conclusion}
\input{Conclusion}

\bibliographystyle{IEEEtran}
\bibliography{ref.bib}












\newpage

\vfill

\end{document}

%% file: abstract.tex
Vision Transformers have achieved remarkable progresses, among which Swin Transformer has demonstrated the tremendous potential of Transformer for vision tasks. It surmounts the key challenge of high computational complexity by performing local self-attention within shifted windows. 
In this work we propose the Semantic-Aware Local-Global Vision Transformer (\emph{SALG}), to further investigate two potential improvements towards Swin Transformer. 
First, unlike Swin Transformer that performs uniform partition to produce equal size of regular windows for local self-attention, our \emph{SALG} performs semantic segmentation in an unsupervised way to explore the underlying semantic priors in the image. As a result, each segmented region can correspond to a semantically meaningful part in the image, potentially leading to more effective features within each of segmented regions. Second, instead of only performing local self-attention within local windows as Swin Transformer does, the proposed \emph{SALG} performs both 1) local intra-region self-attention for learning fine-grained features within each region and 2) global inter-region feature propagation for modeling global dependencies among all regions. Consequently, our model is able to obtain the global view when learning features for each token, which is the essential advantage of Transformer. Owing to the explicit modeling of the semantic priors and the proposed local-global modeling mechanism, our \emph{SALG} is particularly advantageous for small-scale models when the modeling capacity is not sufficient for other models to learn semantics implicitly. Extensive experiments across various vision tasks demonstrates the merit of our model over other vision Transformers, especially in the small-scale modeling scenarios.

%% file: Introduction.tex
\IEEEPARstart{T}{ransformer} has achieved significant success in Natural Language Process (NLP) due to its powerful capability of capturing long-range dependencies in sequence modeling~\cite{vaswani2017attention,devlin2018bert,liu2019roberta}. It has also been adapted to Computer Vision (CV) in recent years and shows promising performance in various vision tasks such as image recognition~\cite{dosovitskiy2020image,touvron2021training,liu2021swin}, object detection~\cite{carion2020end,zhu2020deformable,song2021vidt}, image synthesis~\cite{parmar2018image,esser2021taming,jiang2021transgan} and video classification~\cite{bertasius2021space,arnab2021vivit,patrick2021keeping,fan2021multiscale,wang2022deformable}. Compared with convolutional neural networks (CNNs), an essential merit of Transformer is that it has the global view, namely referring to all tokens, when learning features for each token by the self-attention operation, leading to potentially more effective features than CNNs.

\begin{figure}[!t]
\centering
		\includegraphics[width=1.0\linewidth]{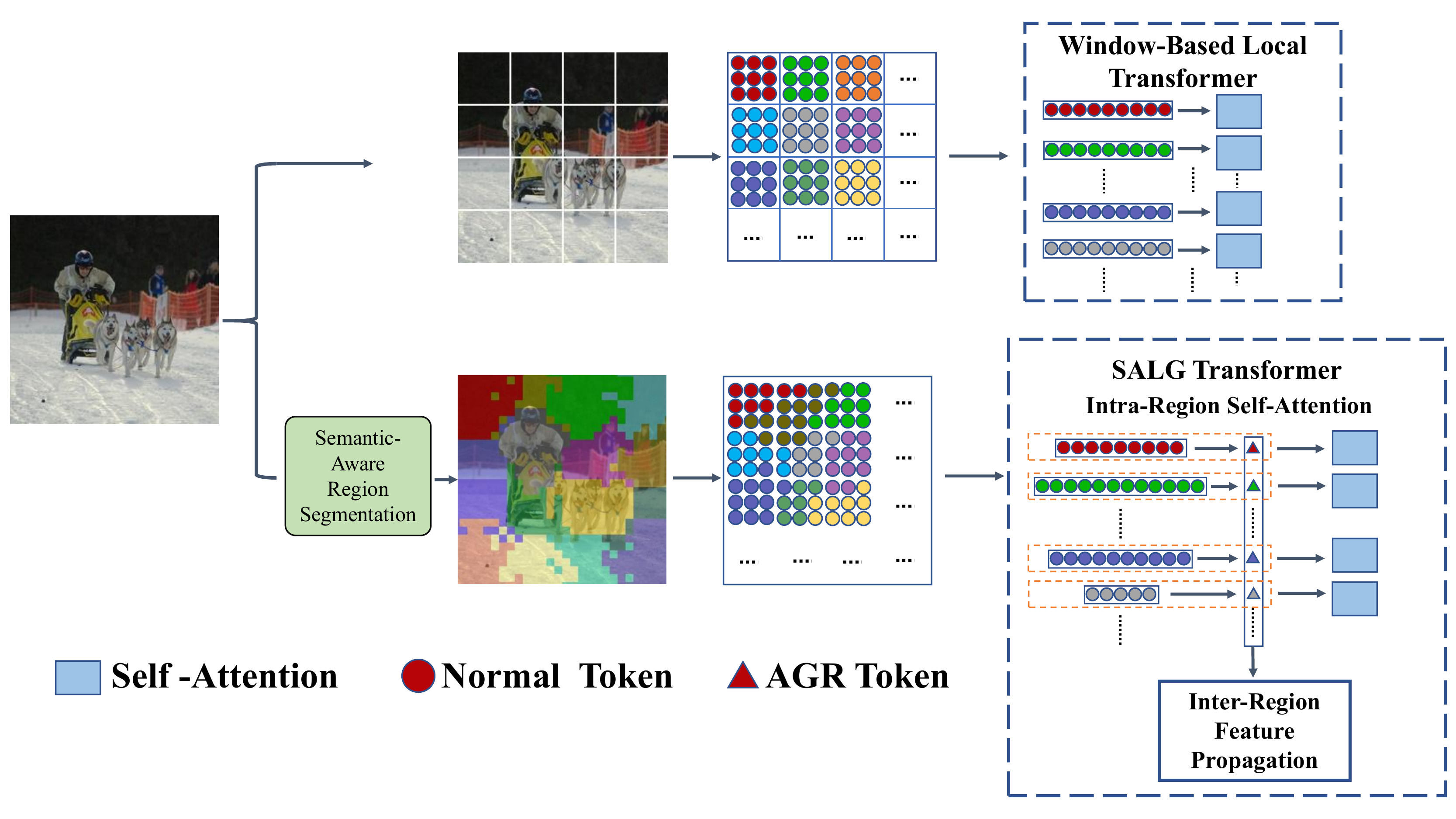}
\centering
\caption{Comparison of Window-based local Transformer and our \emph{SALG} Transformer. 
Unlike typical way of partitioning the feature maps equally, our \emph{SALG} performs unsupervised semantic segmentation to explore the underlying semantic priors in the image. The segmented regions are then processed by \emph{SALG} to learn visual features following the proposed local-global stepwise modeling mechanism: it first performs local intra-region self-attention to learn fine-grained features within reach region, and then models the global dependencies between the aggregation tokens (AGR token) of different regions by global inter-region feature propagation.}
\label{fig:teaser}
\end{figure}

A straightforward way to adapt Transformer to the vision domain, as ViT~\cite{dosovitskiy2020image} does, is to reshape the encoded feature map of an input image into a sequence of pixels and view each pixel as a token. Then Transformer can be directly applied to the sequence for learning features. Nevertheless, the crucial challenge of such method lies in the quadratic computational complexity w.r.t. the image size, which limits its application to high-resolution images. To address this limitation, a typical and also effective solution is to perform self-attention in a local region rather than in a global view~\cite{huang2019ccnet,hu2019local,ramachandran2019stand,wang2020axial,huang2019interlaced,child2019generating,liu2021swin,chu2021twins}, which resembles the local convolutional operation in CNNs. A prominent example is Swin Transformer~\cite{liu2021swin}, which divides the feature map into non-overlapping windows equally and performs self-attention locally within each window. It designs a novel operation mechanism dubbed `shifted windows' to enable interactions between adjacent windows. Besides, the hierarchical Transformer structure is adopted to learn features with multiple scales of receptive field, which makes it particularly effective for dense prediction such as object detection and semantic segmentation.

Despite the great success of Swin Transformer, there remains two potential limitations. First, Swin Transformer partitions the feature map into equal size of regular windows for performing the local self-attention. However, such window segmentation does not take into account the semantic distributions of tokens (pixels) in the feature map. As a result, the segmented windows may contain no semantically meaningful content or randomly mixed content from multiple objects, which has adverse effect on feature learning. A well window segmentation scheme should be performed based on pixel semantics distributions of the feature map so that each of the segmented windows corresponds to either a (part of) object or a patch of background, thereby leading to meaningful features learned by the local self-attention within each window. The second potential limitation of Swin Transformer is that it only performs local self-attention within each window when learning features for each token rather than in a global view, although it performs interactions between adjacent windows and can also obtain expanding receptive field as the increase of attention layers. In this sense it does not inherit perfectly the essential advantage of Transformer, namely learning features in a global view.

To circumvent these two limitations mentioned above, in this work we propose the Semantic-Aware Local-Global Vision Transformer (\emph{SALG}), which is designed as a general-purpose Vision Transformer. Similar to Swin Transformer, our \emph{SALG} Transformer also follows a hierarchical framework consisting of multiple cascaded feature learning stages. Each learning stage comprises two core steps: region segmentation and feature learning. Different from the typical way of equally partitioning the feature maps into regular windows by most existing methods~\cite{liu2021swin,huang2021shuffle,ren2022shunted,fang2022msg}, the proposed \emph{SALG} partitions the feature map into irregular semantic regions in an unsupervised way, which explores the underlying semantic priors in the feature map. As a result, each segmented region can correspond to a semantically meaningful part in the input image such as a (part of) object or a patch of background, as shown in Figure~\ref{fig:teaser}. The segmented semantic regions are then processed by our \emph{SALG} to learn visual features in two steps following the proposed local-global modeling mechanism. First, it performs local intra-region  self-attention to learn fine-grained features for each segmented region. Then \emph{SALG} models the global dependencies between different regions by global inter-region feature propagation. To improve the modeling efficiency, \emph{SALG} learns an extra aggregation token for each region to summarize the features for the entire region. Then the global inter-region feature propagation is implemented by performing self-attention between aggregation tokens of different regions. As a result, our model is able to obtain the global view when learning features for each token and thus retain the essential advantage of Transformer. Meanwhile, our model substantially reduces the computational complexity compared to the typical global self-attention operation of Transformer, benefiting from such local-global stepwise learning mechanism. 

Owing to the explicit modeling of the semantic priors, our model can potentially learn more effective features than the typical Vision Transformers that partition the feature map into equal windows without considering the semantic distributions of tokens, with the comparable model size and computational complexity. Our \emph{SALG} is particularly advantageous for small-scale models when the modeling capacity is not sufficient to learn semantics implicitly. To evaluate the proposed \emph{SALG}, we conduct extensive experiments across various vision tasks including image classification, object detection and instance segmentation. In particular, we build our \emph{SALG} in various scales, including `small', `tiny' and `super-tiny' versions (from large to small in order), and compare them with the corresponding versions of the state-of-the-art vision Transformers with comparable model size.
Experimental results reveal following observations. 1) Our \emph{SALG} largely outperforms other methods on `tiny' and `super-tiny' versions on all three tasks, with larger performance gain in smaller-scale version. Taking the `super-tiny' version as an example, our \emph{SALG} surpasses Swin Transformer by $2.3\%$ in terms of Top-1 accuracy for image classification and $4.7\%$ in terms of box AP for object detection (in 3$\times$ schedule). 2) \emph{SALG} achieves comparable performance with Swin Transformer on `small' version. These results demonstrate the advantages of our model compared with other methods in the small-scale modeling scenarios.

%% file: Related-work.tex
\smallskip\noindent\textbf{Vision Transformer}
Transformer has shown its superior performance in both NLP and vision filed. Different from CNN, whose reception filed is local and limited, vision Transformer models the global information of feature map by self-attention mechanism. As the first vision of Transformer network which achieves comparable capabilities with CNN models, ViT\cite{dosovitskiy2020image} divides the image into 16$\times$16 patches and applies Transformer on the image patch sequence. 
Subsequent works\cite{touvron2021training,han2021transformer,yuan2021tokens} explore improvements to Transformer architecture in vision filed and achieve promising effects. TNT\cite{han2021transformer} proposes to additionally model the relationship of the super-pixel within each image patch to get fine-gained features. T2T-ViT\cite{yuan2021tokens}
proposes a novel tokenization way in T2T module to model the local structure feature of the images. However, these vision Transformer models show their limitations when they come to downstream vision tasks such as detection and segmentation. These dense  prediction vision tasks require higher resolution image inputs and multi-scale features output. Simply applying global attention to these tasks will cause huge amount of computation in Transformer network. Hence it is necessary to design efficient Transformer networks that can handle high resolution image inputs.

\smallskip\noindent\textbf{Efficient Vision Transformer}
Recent works explore two different ways to reduce the complexity of global self-attention in vision Transformer: 1) applying local self-attention instead of global self-attention in Transformer architecture and 2) merging tokens to reduce the number of keys and values interacted with each query token.

For the first strategy, Swin Transformer\cite{liu2021swin} divides the feature map into non-overlap squared windows and then performs local self-attention within these windows. Besides, Swin Transformer exchanges information between different windows by shifting the window partition between successive Transformer layers. Later works such as Shuffle Transformer~\cite{huang2021shuffle} and MSG-Transformer~\cite{fang2022msg} also adopt the window-based local self-attention mechanism, while changing the ways of information exchange between different windows. Shuffle Transformer 
applies spatial shuffle operation to model the cross-window relationship. MSG-Transformer~\cite{fang2022msg} leverages extra messenger tokens to aggregate information of different windows and performs shuffle operation on these messenger tokens.  
Besides, these models all employ hierarchical architecture, which generate multi-scale output feature maps and applied to dense prediction tasks such as detection and segmentation. However, above models using local self-attention rudely divide the feature into regular regions, without considering the semantic information of tokens and the association between them. In contrast, our \emph{SALG} performs semantic segmentation on the feature map in an unsupervised way, resulting in irregular regions. Subsequently, local attention within each region and global attention between regions are applied to extract fine-grained features and model global dependencies, respectively.

In addition, as the representative of the second strategy, PVT\cite{wang2021pyramid} 
proposes a spatial-reduction attention to reduce the number of keys and values participating in self-attention. Thus PVT reduces computational complexity of Transformer and can handle higher resolution input feature maps. However, this strategy is limited in modeling the fine-grained features of small objects as the spatial-reduction operations. Differently, our \emph{SALG} maintains the number of tokens in each layer and learns fine-grained features with local self-attention within each semantic region.

\begin{figure*}[!t]
\centering
\includegraphics[width=\linewidth]{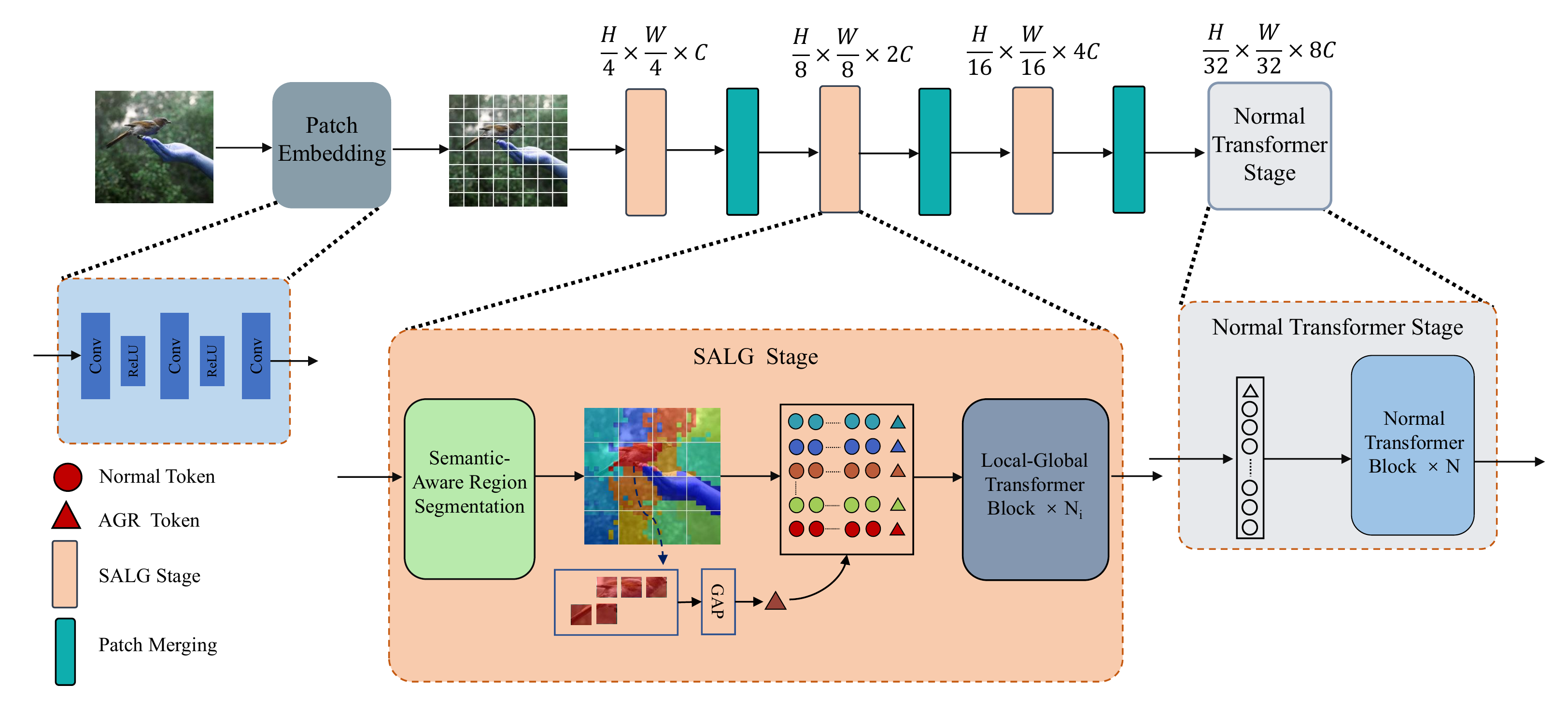}
\caption{ Overall architecture of Semantic-Aware Local-Global (\emph{SALG}) Vision Transformer. The input image is first processed by a patch embedding layer to obtain an initial feature map. Then our \emph{SALG} learns visual features by four cascaded learning stages including three specialized \emph{SALG} stages and one normal Transformer stage. Each \emph{SALG} stage first performs semantic-aware region segmentation in an unsupervised way to explore the underlying semantic priors. Then the \emph{SALG} stage learns features by the designed Local-Global Transformer block, which performs both local intra-region self-attention within each semantic region and global inter-region feature propagation to model dependencies between regions.}
\label{fig:fig_sim}
\end{figure*}

\smallskip\noindent\textbf{Deformable Attention}
Deformable convolution\cite{dai2017deformable} predicts spatial positions to apply convolution filters according to the input semantic information. Similarly, deformable attention models\cite{xia2022vision,zhu2020deformable} also learn to focus on some limited relevant tokens in a data-dependant way. Deformable DETR\cite{zhu2020deformable} accelerates the convergence of DETR\cite{carion2020end} by selecting a small number of keys and values for each query token. DAT\cite{xia2022vision} proposes a deformable attention block, which takes the feature map as input and predicts location bias to obtain the global informative regions on the feature map. As an effective video transformer model, DVT \cite{wang2022deformable} proposes to predict $N$ most salient video patches for each query token according to its semantic information and estimated motion clues. With a small number of $N$, DVT decreases the computation complexity and achieves higher accuracy than those models performing global attention.

However, while determining the target key and value pairs, these models only refer to the query token, without considering their semantic relationships. Moreover, they would cause the discontinuity when the determined key and value pairs are scattered. 
In contrast, our \emph{SALG} divides the feature map into continuous regions for attention interaction, through explicitly comparing the semantic similarities between different tokens in local windows. 

%% file: Method.tex

Our Semantic-Aware Local-Global Vision Transformer (\emph{SALG} Transformer) is designed as a general-purpose Transformer for learning visual features, which can be further utilized in various downstream tasks in Computer Vision. Given an input image, the proposed \emph{SALG} Transformer first partitions it into semantic regions in an unsupervised way, which explores the underlying semantic priors in the image. Then \emph{SALG} Transformer employs the designed Local-Global Transformer block iteratively to perform both local intra-region self-attention within each semantic region and global inter-region feature propagation between regions. Consequently, our \emph{SALG} Transformer is able to not only perform fine-grained feature learning locally for each region which can substantially reduce the computational complexity than the global learning manner of conventional Transformer, but also model global dependencies at the region level efficiently, thereby learning effective visual features for downstream tasks.

In this section we will first present the overall framework of the proposed \emph{SALG} Transformer. Then we will elaborate on the unsupervised semantic-aware region segmentation and the designed Local-Global Transformer block in our \emph{SALG} Transformer, respectively.

\subsection{Overall Framework}
\label{sec:3.1}
Figure~\ref{fig:fig_sim} illustrates the overall architecture of our \emph{SALG} Transformer. It learns the visual representations for an input image in an iterative manner by four cascaded feature learning stages, including three specialized \emph{SALG} stages consisting of the designed region segmentation module and Local-Global Transformer block, and one normal stage based on typical Transformer block. 

Given an input image of size $H \times W \times 3$, the proposed \emph{SALG} Transformer first employs a patch embedding layer to project the input image to a feature map of size $\frac{H}{4} \times \frac{W}{4} \times C$, where $C$ is the feature dimension. Recent research~\cite{wang2021pyramid,wang2022scaled,ren2022shunted} has shown that using convolutional operations to perform feature transformation with overlapping of receptive fields between tokens rather than splitting the input image into non-overlapping patches (as ViT~\cite{dosovitskiy2020image} and Swin Transformer do) can obtain higher-quality tokens. Specifically, we use three convolutional layers in the patch embedding layer, among which the stride of the first two convolutional layers are set to 2 to downsample the input image into the size of $\frac{H}{4} \times \frac{W}{4}$. The third convolutional layer with stride 1 is used for feature transformation.

The obtained transformed feature map is then fed into three specialized \emph{SALG} stages and one normal stages sequentially for feature learning. Each \emph{SALG} stage comprises two steps: region segmentation and feature learning. The feature map is partitioned into semantic regions by the designed Semantic-Aware Region Segmentation module in an unsupervised way. Different from the typical way of equally partitioning the feature map into regular windows by most existing methods~\cite{liu2021swin,chen2021regionvit,chu2021twins,vaswani2021scaling,huang2021shuffle,fang2022msg}, our \emph{SALG} Transformer aims to segment the feature map into semantic regions that could be irregular and thus explore the underlying semantic priors. For instance, as shown in Figure~\ref{fig:fig_sim}, the pixels (tokens) from a same object or a same patch of background tend to be partitioned into the same region.

The segmented regions are then processed by the Local-Global Transformer block to learn visual representations. In particular, it first performs local intra-region self-attention for learning fine-grained features for each segmented region. Then the Local-Global Transformer block models the global dependencies between different regions via global inter-region feature propagation. As a result, our model is able to obtain the global view when refining features for each token, which is the essential advantage of Transformer. Meanwhile, our model substantially reduces the computational complexity compared to the typical global self-attention operation of Transformer, benefiting from such Local-Global stepwise learning mechanism.

Similar to Swin Transformer, we adopt the patch merging operation, implemented by a linear transformation layer, to gradually reduce the size of feature maps between stages and thus learn a hierarchical representation. Three Patch Merging modules between four learning stages compress the feature map from $\frac{H}{4} \times \frac{W}{4}$ to $\frac{H}{32} \times \frac{W}{32}$, with increasing feature dimensions (channels) for each pixel. The obtained hierarchical representations can be further used for downstream tasks including image-level prediction like image classification and pixel-level prediction such as object detection or semantic segmentation.

\begin{figure*}[!t]
\centering
\includegraphics[width=\linewidth]{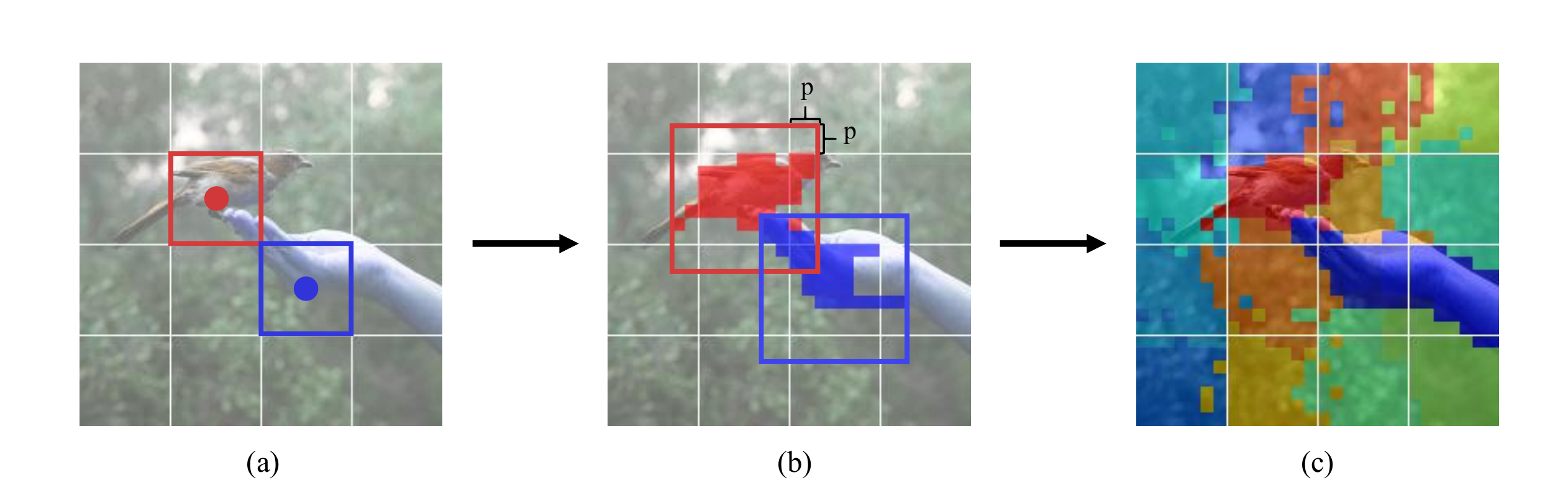}
\caption{Unsupervised semantic-aware region segmentation. (a) Our model first splits the feature map into non-overlapping regular windows uniformly and calculate the average features for each window to approximate the mean representations for semantic regions to be segmented, dubbed `region means'. (b) We define the coverage area for each region mean by expanding its window with $p$ pixels on all four sides. (c) A token (pixel) is assigned to the region mean with maximal Cosine similarity among all region means covering the token.}
\label{fig:fig_seg}
\end{figure*}

\begin{figure*}[!t]
\centering
\includegraphics[width=\linewidth]{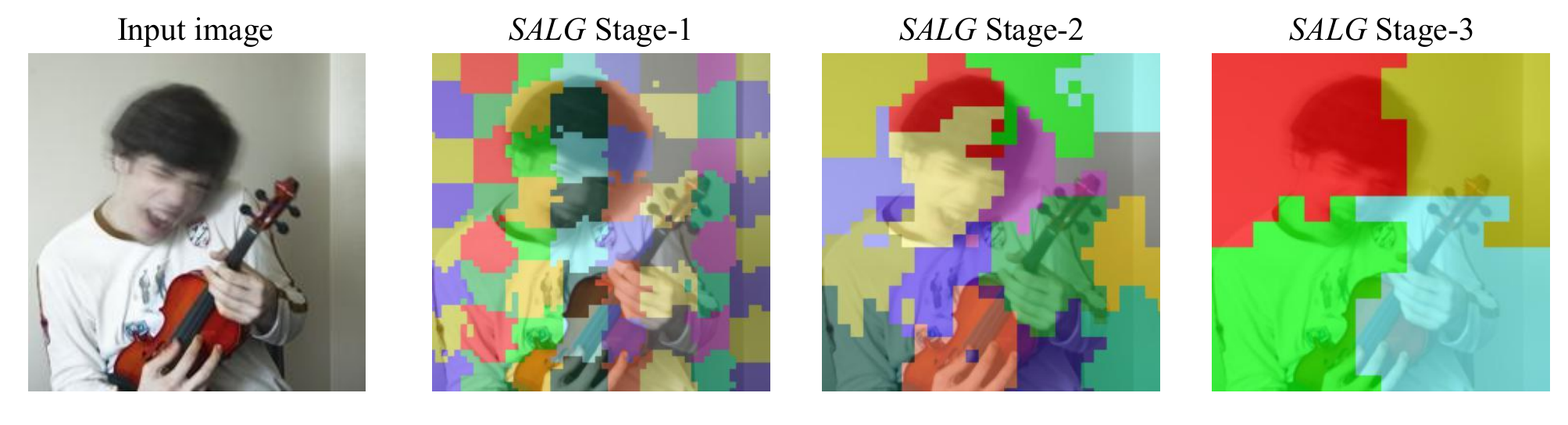}
\caption{Visualization of hierarchical region segmentation by our \emph{SALG} Transformer for a randomly selected sample from Imagenet-1K validation set. The semantic segmentation results in three \emph{SALG} stages are presented. The deeper stages have less segmented regions which correspond to larger receptive fields in the input image.}
\label{fig:hie_segm}
\end{figure*}

\subsection{Unsupervised Semantic-Aware Region Segmentation}
\label{sec:3.2}
We perform semantic-aware region segmentation in an unsupervised way, which does not require the pixel-level annotation. An intuitive way is to perform token (pixel) clustering, for which K-means clustering algorithm is a straightforward method. However, K-means is inefficient due to its iteratively refining strategy. Moreover, performing k-means over the entire feature map ignores the positional continuity between tokens and tends to produce a lot of small disconnected regions belonging to a same cluster. To circumvent above two issues, we adapt K-means algorithm to our Semantic-Aware Region Segmentation module in both core steps: calculation of mean representations for clusters (regions) and token assignment for segmentation. Both steps only need to be conducted once in our Semantic-Aware Region Segmentation module, which is quite efficient.

\smallskip\noindent\textbf{Estimation of mean representations for regions.}
Instead of iteratively updating the cluster means in K-means algorithm, we estimate $K$ mean representations for total $K$ regions (clusters) in the feature map directly. Specifically, we divide the feature map into $n \times n$ non-overlapping rectangle windows uniformly, where $n \times n = K$, then we calculate the average features over all tokens for each window to approximate mean representations of $K$ regions, as shown in Figure~\ref{fig:fig_seg}(a). Estimating region means in such a way leads to two potential merits: 1) the positional continuity is taken into account since each region mean is estimated from a local window of tokens instead of scattered tokens; 2) all region means can be calculated in parallel to achieve high efficiency. 

\smallskip\noindent\textbf{Token assignment for region segmentation.} Similar to K-means algorithm, we assign each token to the region mean with the nearest semantic distance. To preserve the positional continuity, we only assign a token to its nearby region means. Formally, for each region mean, we define its coverage area by expanding its window with $p$ pixels on all four sides, resulting in an area of $(\frac{H'}{n}+2p) \times (\frac{W'}{n}+2p)$. Here $H'$ and $W'$ are the height and width of the feature map being processed, respectively. Consequently, a token can fall into the coverage area of more than one region means due to the overlapping between coverage areas of regions means, as shown in Figure~\ref{fig:fig_seg}(b). Then we calculate the Cosine similarity in the feature space between a token located at $<u,v>$ to all region means covering it, denoted as a set of region means $\mathcal{S}$, and select the region mean $r_{<u,v>}$ with the maximal similarity for assignment:
\begin{equation}
\begin{split}
& s_{<u,v>}^i = \frac{\mathbf{m}_i \cdot \mathbf{F}_{u,v}}{ \Vert \mathbf{m}_i \Vert \Vert \mathbf{F}_{u,v} \Vert}, \qquad \forall i \in \mathcal{S},\\
& r_{<u,v>} = \text{argmax}_{i}{s_{<u,v>}^i}.
\end{split}
\label{eqn:token_assign}
\end{equation}
Herein, $s_{<u,v>}^i$ is the Cosine similarity between the representation $\mathbf{m}_i$ of the $i$-th region mean and the token feature $\mathbf{F}_{u,v}$ located at $<u,v>$. In this way, the whole feature map is segmented into semantic regions in an unsupervised way while basically preserving the positional continuity, as illustrated in Figure~\ref{fig:fig_seg}(c). 

\smallskip\noindent\textbf{Hierarchical region segmentation.} We perform region segmentation in each of the  three \emph{SALG} stages while keeping the window size of region means consistent across different stages. Thus, deeper stages have less regions, each region corresponding to a larger receptive field in the input image, as shown in Figure~\ref{fig:hie_segm} . As a result, we obtain a hierarchical region segmentation, which enables our model to capture the object semantics in  different scales. It is particularly advantageous to tasks of pixel-level predictions like object detection or semantic segmentation, which is experimentally validated in Section~\ref{sec:4.2}.

\smallskip\noindent\textbf{Seamless integration into \emph{SALG} Transformer.} Besides the computational efficiency, another prominent advantage of the proposed Semantic-Aware Region Segmentation module is that it can be seamlessly integrated into the our \emph{SALG} Transformer, which allows for both the end-to-end inference and optimization for the whole model.

\subsection{Local-Global Transformer Block}
\label{sec:3.3}
\begin{figure}[!t]
\centering
\includegraphics[width=1.0\linewidth]{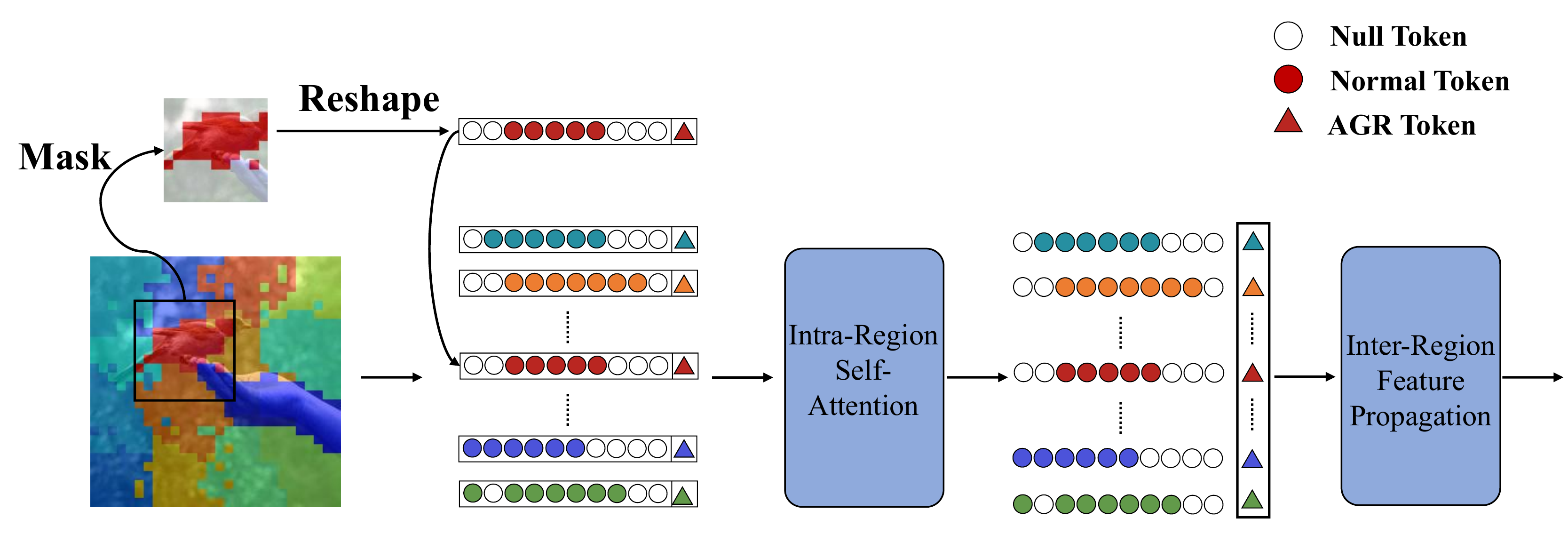}
\caption{Illustration of Local-Global Transformer block, which consists of local intra-region self-attention and global inter-region feature propagation. The local intra-region self-attention is performed in the coverage area of equal size for each semantic region while masking out the tokens (denoted as `null token' in the figure) that do not belong to current region for parallel computing. The inter-region feature propagation is performed on the ARG tokens which aggregate representative features for each region.}
\label{fig:local-global}
\end{figure}

We design the Local-Global Transformer block of our \emph{SALG} Transformer to learn visual features from segmented regions in two successive steps as shown in Figure~\ref{fig:local-global}: local feature learning within each region and global feature propagation between regions, which allows \emph{SALG} Transformer to learn features efficiently in a global view.

\smallskip\noindent\textbf{Intra-Region Self-Attention for learning local object-wise semantics.}
The segmented semantic regions generally correspond to either a (part of) object or a patch of background. We first perform fine-grained feature learning within each region to extract object-wise semantics. To this end, we perform intra-region self-attention operations for each region individually. Different from the typical methods that partition the feature map equally, our \emph{SALG} Transformer obtains irregular regions of unequal size from the Semantic-Aware Region Segmentation module. To enable parallel intra-region self-attention for all regions, for each region we implement local self-attention operation in the coverage area of the corresponding region mean while masking out the tokens in the coverage area that do not belong to current region (denoted as null tokens in the figure). Such implementation is based on three observations: 1) all region means have the same size of coverage area, i.e., $(\frac{H'}{n}+2p) \times (\frac{W'}{n}+2p)$, 2) each segmented region is bounded by the coverage area of its region mean, 3) the spatial structure of all tokens in the entire feature map is preserved, which allows for the straightforward incorporation of the relative position encoding.

For a segmented region, apart from the normal tokens in the coverage area of its region mean, we also add a learnable token, termed as Aggregation (AGR) Token shown in Figure~\ref{fig:local-global}, to learn the aggregated representation for the entire region. The AGR token is responsible for feature interaction between regions during the inter-region feature propagation and also providing the global information of the whole feature map in the operation of intra-region self-attention. As a result, our \emph{SALG} Transformer is able to obtain the global view when refining features for each token locally via the intra-region self-attention. Formally, denoting the feature map of a segmented region $r$ as $\mathbf{F}_r$, the intra-region self-attention is modeled as:
\begin{equation}
\begin{split}
    &\mathbf{F}'_r = \mathbf{F}_r \odot \mathbf{M}_r, \\
    &\mathbf{x}_r = \text{Concat} (\mathbf{t}^\text{AGR}_r, \text{Flatten}(\mathbf{F}'_r)), \\
    &\mathbf{x}_{r} = \text{Local-MSA}(\mathrm{LN}(\mathbf{x}_{r}))+\mathbf{x}_{r},\\
    &\mathbf{x}_{r} = \text{MLP}(\text{LN}(\mathbf{x}_{r}))+\mathbf{x}_{r}.
\end{split}
\end{equation}
Here $\mathbf{M}_r$ denotes the mask to indicate whether a token in a coverage area belong to current region (1 for true and 0 for false) and $\mathbf{t}^\text{AGR}_r$ is the feature for AGR token for $r$. `Local-MSA' refers to the local operation of Multi-head Self-Attention within the coverage area of the region mean for $r$. The operation `Flatten' reshapes the coverage area $\mathbf{F}'_r$ into an array of tokens. As other Vision Transformers~\cite{liu2021swin} do, the relative position encoding~\cite{shaw2018self} is used to provide relative spatial information during feature learning.

\smallskip\noindent\textbf{Inter-Region Feature Propagation for modeling global dependencies.}
Our \emph{SALG} Transformer performs inter-region feature propagation to model the global dependencies between different regions. Specifically, we perform feature interaction between AGR tokens of different regions to improve the computational efficiency, as illustrated in Figure~\ref{fig:local-global}. The refined AGR tokens after current inter-region propagation are then fed to the intra-region self-attention in the next layer, which provides global information of the whole feature map for local feature learning for each tokens. Such local-global stepwise learning mechanism substantially reduce the computational complexity while preserving the essential advantage of Transformer, namely learning features for each token in a global view.

We investigate two optional implementations for inter-region feature propagation. The first way is to apply Multi-head Self-Attention (MSA) to total $n^2$ AGR tokens to propagate features, which is modeled by:
\begin{equation}
    \begin{split}
    &\mathbf{T}^\text{AGR} = \text{Concat}(\mathbf{t}^\text{AGR}_1, \dots, \mathbf{t}^\text{AGR}_{n^2}), \\
    &\mathbf{T}^\text{AGR} = \text{MSA}(\mathrm{LN}(\mathbf{T}^\text{AGR})+\mathbf{T}^\text{AGR},\\
    &\mathbf{T}^\text{AGR} = \text{MLP}(\text{LN}(\mathbf{T}^\text{AGR}))+\mathbf{T}^\text{AGR}.
    \end{split}
    \label{eqn:agr-msa}
\end{equation}
While such way of inter-region feature propagation is able to achieve deep feature interaction between AGR tokens of different regions, a potential limitation is that it introduces extra parameters and thus increases the computational complexity to some degree. A more simple and efficient solution is to perform average fusion on all AGR features via an averaging pooling (AP) operation:
\begin{equation}
\begin{split}
&\mathbf{T}^\text{AGR} = \text{Concat}(\mathbf{t}^\text{AGR}_1, \dots, \mathbf{t}^\text{AGR}_{n^2}), \\
&\mathbf{T}^\text{AGR} = \text{AP}(\mathbf{T}^\text{AGR}),
\end{split}
\label{eqn:agr-avg}
\end{equation}
where the obtained $\mathbf{T}^\text{AGR}$ is assigned to each of all AGR tokens. In this case, all AGR tokens are treated equally and no fine-grained distinctions between regions are modeled during the inter-region feature propagation. We conduct experimental comparison for ablation study in Section~\ref{sec:4.3}.

\subsection{Different Scales of Architecture Variants}
\label{sec:3.4}
Our \emph{SALG} Transformer first performs semantic region segmentation to explore the semantic priors, and then learns fine-grained features locally within each segmented region and models the inter-region dependencies globally. Compared to typical Vision Transformers that partition the feature map into equal windows without considering the semantic distributions of tokens, our model can potentially learn more effective features with the comparable model size and computational complexity due to the explicit modeling of the semantic priors. Our \emph{SALG} is particularly advantageous for small-scale models when the modeling capacity is not sufficient to learn semantics implicitly. 

To have a fair and extensive comparison with the existing Vision Transformers, especially in the small-scale modeling scenarios, we build our \emph{SALG} in various scales, to have comparable model size and computational complexity with the corresponding versions of DeiT and Swin Transformer. To be specific, we build tiny (\emph{SALG}-T) and small (\emph{SALG}-S) versions, which correspond to the tiny and small versions of Swin, as well as the small and base versions of DeiT respectively. Besides, we also introduce an extra super-tiny version (\emph{SALG}-ST), which is even smaller than the tiny version, to investigate the effectiveness of our model in extremely small scale. We also build the corresponding super-tiny version for both DeiT and Swin accordingly for comparison. Note that we do not consider the `base' and `large' versions of different models since we focuses on small-scale modeling scenarios. We adjust the feature dimension $C$ and the numbers of Local-Global Transformer block in each stage to achieve different scales of our model:
\begin{itemize}
    \item \emph{SALG}-ST: C = 48, block numbers = \{2, 2, 2, 2\};
    \item \emph{SALG}-T: C = 96, block numbers = \{2, 2, 6, 2\};
    \item \emph{SALG}-S: C = 96, block numbers = \{2, 2, 18, 2\}.
\end{itemize}

%% file: Experiment.tex
\begin{figure*}[!t]
\centering
\includegraphics[width=\linewidth]{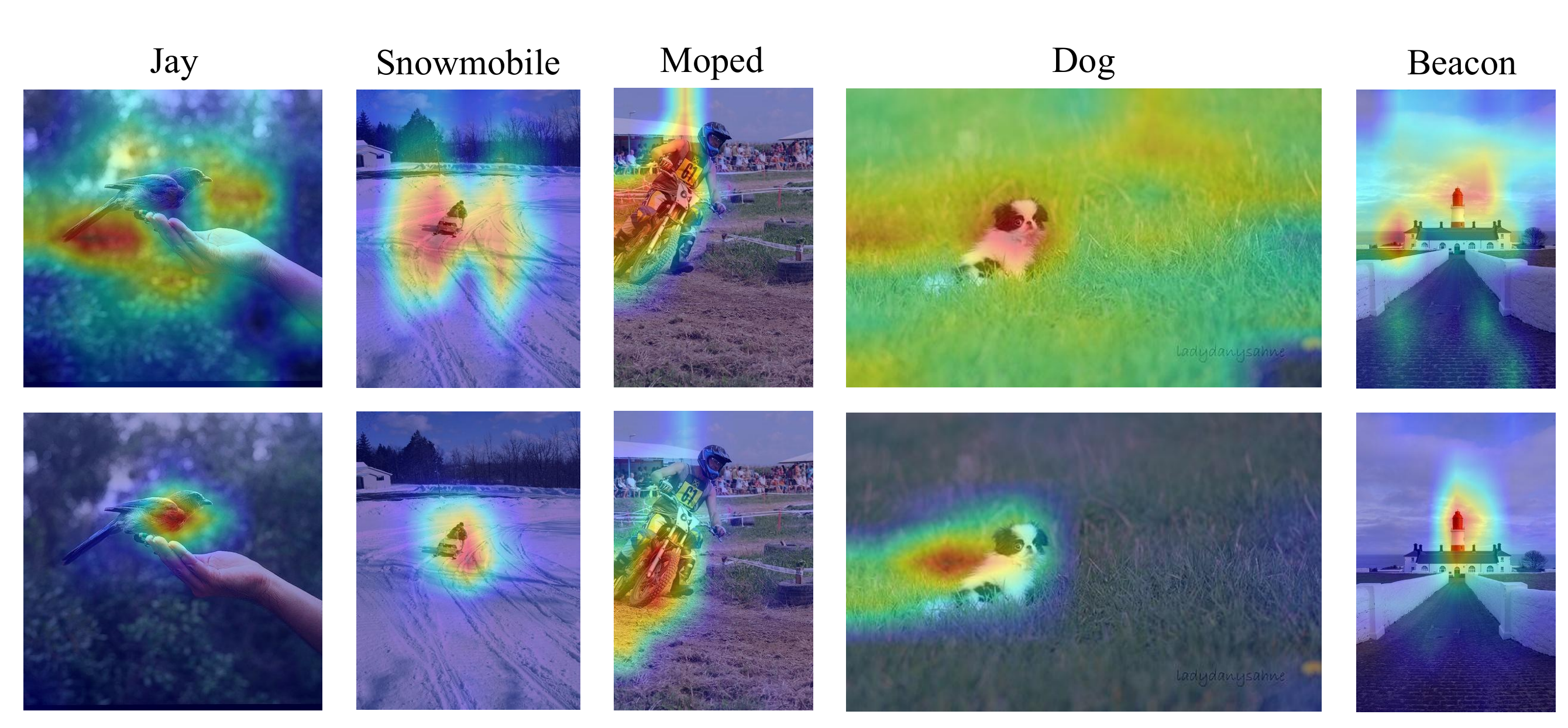}
\caption{Visualization of class activation maps (CAM) of Swin Transformer (the first row) and our \emph{SALG} Transformer (the second row) in the `super-tiny' modeling scale on four randomly selected images from Imagenet-1K validation set. These maps highlight the salient regions relevant to the target classes, which manifests that our model is able to capture the related object to the class label more precisely than Swin Transformer.
}
\label{fig:figure5}
\end{figure*}

To evaluate the generalization and effectiveness of our \emph{SALG} Transformer model in various task of Computer Version, we first apply our model to ImageNet-1K~\cite{deng2009imagenet} dataset for classification in Section~\ref{sec:4.1}. Then we evaluate our model on COCO 2017~\cite{lin2014microsoft} for object detection and  instance segmentation in Section~\ref{sec:4.2}. We also perform a series of ablation studies and analysis to validate the effectiveness of different components of our model in Section~\ref{sec:4.3}.

\subsection{Image Classification}
\label{sec:4.1}
\smallskip\noindent\textbf{Dataset and Experimental Setups.}
We conduct experiments to compare our model with state-of-the-art methods for classification on ImageNet-1K dataset~\cite{deng2009imagenet}. ImageNet-1K contains 1.28 million training images and 50K validation images over 1K classes. 
In particular, we compare our model with Swin Transformer which has shown its superior performance on various vision tasks. Besides, we also compare our model with other Transformer-based backbone models including DeiT~\cite{touvron2021training}, T2T~\cite{yuan2021tokens}, PVT~\cite{wang2021pyramid} and TNT~\cite{han2021transformer}.
In experiments, we follow most of training settings in Swin Transformer~\cite{liu2021swin}. Specifically, the training process takes 300 epochs and 20 warm-up epochs. We adopt AdamW~\cite{kingma2014adam} optimizer with weight decay of 0.05 and a cosine annealing decay learning rate schedule~\cite{loshchilov2016sgdr}. The initial learning rate is set to 0.0005 and the batch size is set to 128 due to the limitation of memory. Moreover, we adopt Mixup~\cite{yun2019cutmix}, Random Erasing~\cite{zhong2020random} and label smoothing~\cite{szegedy2016rethinking} in our implementation. 

\begin{table}[!t]
\caption{Comparisons between different models for image classification  on ImageNet-1K.}
\centering
\label{tab:table1}
\resizebox{0.9\linewidth}{!}{
\begin{tabular}{c|l|c|c}
\toprule
\textbf{Model Scale} & \textbf{Methods}  & \textbf{Params} & \textbf{Top-1(\%)} \\
\midrule
\multirow{4}{*}{Super-tiny} & ResNet-18\cite{he2016deep} & 11.7M  & 69.8 \\
&DeiT-T\cite{touvron2021training} & 5.7M  & 72.2 \\
&Swin-ST & 5.6M  &73.6 \\
&SALG-ST (ours) & 6.5M  & \textbf{75.9} \\
\midrule
\multirow{7}{*}{Tiny} &ResNet-50\cite{he2016deep} & 25M  & 76.2 \\
&DeiT-S\cite{touvron2021training} & 22M  &79.8 \\
&T2T-$\rm ViT_{t}$-14 \cite{yuan2021tokens} & 22M  & 80.7\\
&PVT-S \cite{wang2021pyramid} & 25M  & 79.8 \\
&TNT-S \cite{han2021transformer} & 24M  & 81.3 \\
&Swin-T \cite{liu2021swin} & 29M  &81.3 \\
&SALG-T (ours) & 32M  & \textbf{82.1} \\
\midrule
\multirow{7}{*}{Small} &DeiT-B\cite{touvron2021training} & 86M  &81.8 \\
&T2T-$\rm ViT_{t}$-19 \cite{yuan2021tokens} & 39M  & 81.4\\
&T2T-$\rm ViT_{t}$-24 \cite{yuan2021tokens} & 64M  & 82.2\\
&PVT-L \cite{wang2021pyramid} & 61M  & 81.7 \\
&TNT-B \cite{han2021transformer} & 66M  & 82.8 \\
&Swin-S \cite{liu2021swin} & 50M  & \textbf{83.0} \\
&SALG-S (ours) & 53M  &\textbf{83.0} \\
\bottomrule
\end{tabular}
}
\end{table}

Besides the specific model setting of different architecture variants described in Section~\ref{sec:3.4}, following Swin Transformer~\cite{liu2021swin}, we take common model configuration that the feature dimension of each head is set to 32 and the expansion rate in MLP layers is set to 4. In the process of semantic-aware region segmentation, the size of non-overlapping patches is set to $7\times7$, and the numbers of tokens padded on four sides of windows in coverage area are set to 1, 2 and 1 respectively in the three Local-Global Transformer blocks.

\smallskip\noindent\textbf{Results on Image Classification.}
We evaluate our \emph{SALG} Transformer network with different scales on ImageNet-1K and report the Top-1 accuracy on validation set in Table~\ref{tab:table1}. For each variant of our \emph{SALG} Transformer, we compare it with other Conv-based and Transformer-based backbones with comparable model size and computational complexity. Specifically, to evaluate the performance of our \emph{SALG} Transformer in the small-scale modeling scenarios, we trained another Swin Transformer variant, called Swin-ST, which has the same feature dimension and number of layers in each stage as in our super-tiny version of \emph{SALG}.
Besides, for a fair comparison, we adopt averaging
pooling (AP) operation for inter-region feature propagation in Table~\ref{tab:table1}, thus our model has similar model size and computational complexity as other models. The performance of our model with the inter-region feature propagation of multi-head self-attention on Imagenet-1K is shown in Table~\ref{tab:table4}.\\
\indent As shown in Table~\ref{tab:table1}, our models achieve superior performance compared to other Transformer backbone models with similar parameters. In particular, our \emph{SALG} Transformer shows more remarkable advantages in smaller-scale scenarios when the modeling capacity is not sufficient to learn semantics implicitly by other models. For instance, with only 0.8M and 0.9M increments of parameters, \emph{SALG}-ST achieves 3.7$\%$ and 2.3$\%$ Top-1 accuracy gain over DeiT-T and Swin-ST respectively. Meanwhile, our \emph{SALG} Transformer outperforms other popular vision Transformer models such as T2T~\cite{yuan2021tokens}, PVT~\cite{wang2021pyramid} and TNT~\cite{han2021transformer} with comparable model sizes in all scenarios.

\smallskip\noindent\textbf{Qualitative evaluation.}
To obtain more insight into the comparison of the salient regions that different models pay attention to when learning features, we perform qualitative comparison between our \emph{SALG} Transformer and Swin Transformer~\cite{liu2021swin} in the `super-tiny' modeling scale. We randomly choose four samples from the validation set of Imagenet-1K~\cite{deng2009imagenet} and visualize the class activation maps (CAM)~\cite{zhou2016learning} of the target classes in Figure~\ref{fig:figure5}. Comparing the highlighted regions relevant to the target classes, we observe that our \emph{SALG} Transformer is able to focus more precisely on the foreground objects relating to the class label compared to Swin Transformer~\cite{liu2021swin}. These results reveal the advantages of explicitly exploring semantic priors and the local-global attention mechanism in our \emph{SALG} Transformer.

\begin{table*}[t]
\caption{Comparison between different models for object detection and instance segmentation on COCO 2017.}
\label{tab:table2}
\centering
\resizebox{\linewidth}{!}{
\begin{tabular}{c|l|c|ccc|ccc|ccc|ccc}
\toprule
\multirow{2}{*}{\textbf{Model Scale}}&\multirow{2}{*}{\textbf{BackBone}} & \textbf{Params} & \multicolumn{6}{c|}{\textbf{1$\times$ schedule}} & \multicolumn{6}{c}{\textbf{3$\times$ schedule}} \\
& & (M) & $AP^b$ & $AP^b_{50}$ & $AP^b_{75}$ & $AP^m$ & $AP^m_{50}$& $AP^m_{75}$& $AP^b$ & $AP^b_{50}$ & $AP^b_{75}$ & $AP^m$ & $AP^m_{50}$& $AP^m_{75}$\\
\midrule

\multirow{2}{*}{Super-tiny}&Swin-ST & 25 & 34.4 & 56.4 & 36.6 & 32.4 & 53.4 & 33.9 & 36.3 & 57.6 & 39.0 & 36.3 & 57.6 & 39.0 \\
&SALG-ST (ours) & 26 & \textbf{37.6} & \textbf{60.5} & \textbf{40.3} & \textbf{35.3} & \textbf{57.6} & \textbf{37.3} & \textbf{41.0} & \textbf{63.8} & \textbf{44.5} & \textbf{38.2} & \textbf{60.7} & \textbf{40.8} \\

\midrule
\multirow{4}{*}{Tiny}&Res50 & 44 & 38.0 & 58.6 & 41.4 & 34.4 & 55.1 & 36.7  &41.0 & 61.7 & 44.9 & 37.1 & 58.4 & 40.1\\
&PVT-S & 44 & 40.4 & 62.9 & 43.8 & 37.8 & 60.1 & 40.3 & 43.0 & 65.3 & 46.9 & 39.9 & 62.5 & 42.8 \\
&Swin-T & 48 & 42.2 & 64.6 & 46.2 & 39.1 & 61.6 & 42.0 & 46.0 & 68.2 & 50.2 & 41.6 & 65.1 & 44.8 \\
&SALG-T (ours) & 52  & \textbf{43.7} & \textbf{67.3} & \textbf{47.8} & \textbf{40.0} & \textbf{63.8} & \textbf{42.8} & \textbf{46.5} & \textbf{68.9} & \textbf{50.8} & \textbf{42.2} & \textbf{65.9} & \textbf{45.3} \\

\midrule
\multirow{4}{*}{Small}&Res101 & 63 & 40.4 & 61.1 & 44.2 & 36.4 & 57.7 & 38.8  &42.8 & 63.2 & 47.1 & 38.5 & 60.1 & 41.3\\
&PVT-M & 64 & 42.0 & 64.4 & 45.6 & 39.0 & 61.6 & 42.1 & 44.2 & 66.0 & 48.2 & 40.5 & 63.1 & 43.5 \\
&Swin-S & 69 & \textbf{44.8} & 66.6 & 48.9 & \textbf{40.9} & 63.6 & \textbf{44.2} &\textbf{48.5} & \textbf{70.2} & \textbf{53.5} & \textbf{43.3} & \textbf{67.3} & \textbf{46.6} \\
&SALG-S (ours) & 73 & \textbf{44.8} & \textbf{68.0} & \textbf{49.1} & 40.4 & \textbf{64.2} & 43.1 & 46.5 & 68.8 & 51.1 & 41.7 & 65.5 & 44.8\\
\bottomrule
\end{tabular}
}
\end{table*}


\subsection{Object Detection and Instance Segmentation}
\label{sec:4.2}
\smallskip\noindent\textbf{Dataset and Evaluation Metrics.}
We perform object detection and instance segmentation on COCO 2017 dataset, which contains 118K training images and 5K validation images, to evaluate the effectiveness of our \emph{SALG} Transformer as the backbone model in downstream tasks. Following previous works~\cite{liu2021swin,fang2022msg,wang2021pyramid,ren2022shunted}, we report the bbox mAP and mask mAP, noted as $AP^b$ and $AP^m$ respectively, over multiple thresholds as the evaluation metrics in our experiments.

\begin{figure*}[!t]
\centering
\includegraphics[width=\linewidth]{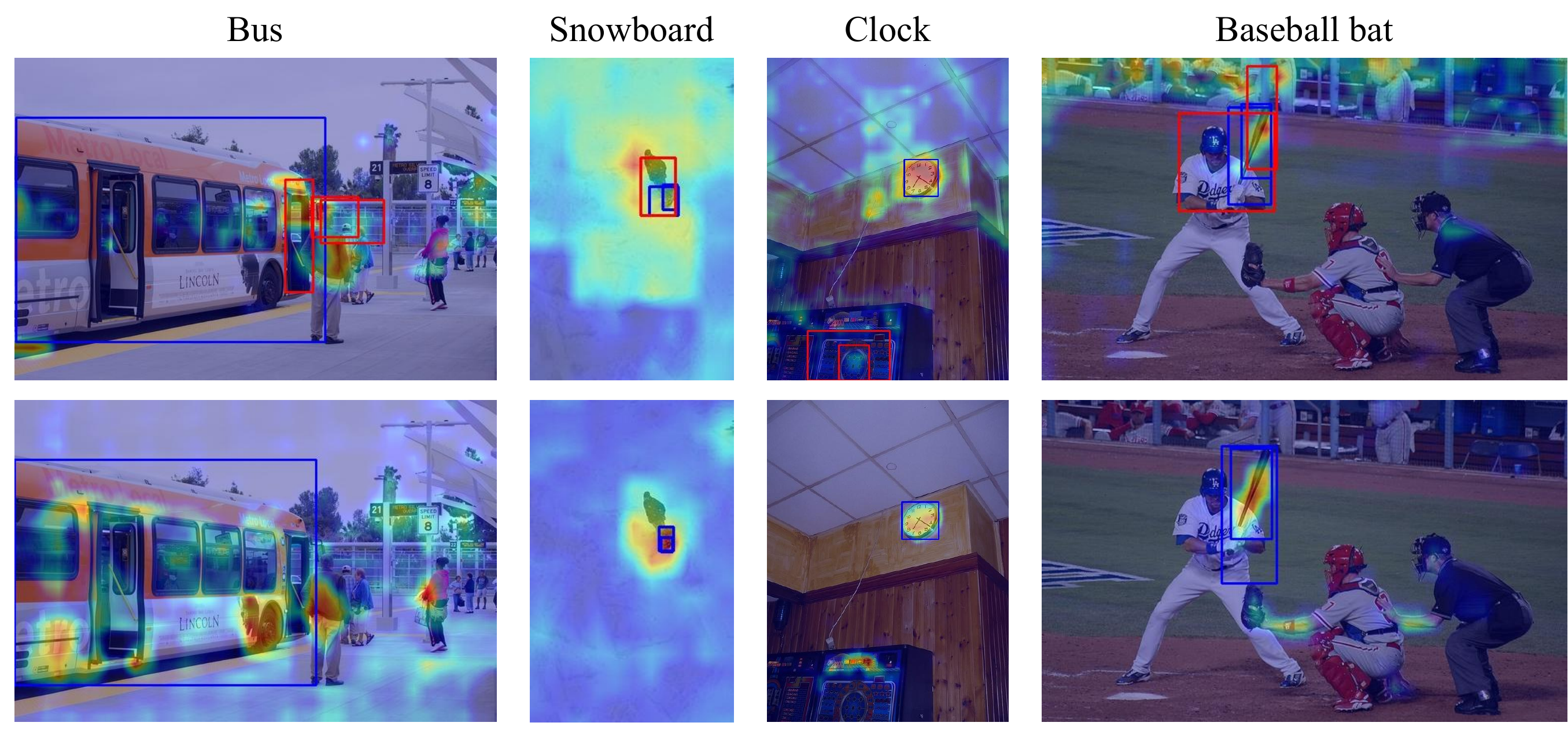}
\caption{Visualization of class activation maps (CAM) of Swin Transformer (in the first row) and our SALG Transformer (in the second row) in the `super-tiny' modeling scale on four randomly selected images from COCO-2017 validation set. Blue bounding boxes indicate correct predictions for object detection while red bounding boxes are wrong predictions. The maps highlight the salient regions for each model, which show that our model can capture the target objects more precisely than Swin Transformer, leading to more precise object detection. By contrast, Swin Transformer makes several wrong predictions indicated by red bounding boxes.}
\label{fig:figure6}
\end{figure*}

\smallskip\noindent\textbf{Training and Evaluation.}
For the tasks of detection and segmentation, we first pretrain our model on ImageNet-1K for 300 epochs. Then we adopt Mask R-CNN\cite{he2017mask} framework and apply our \emph{SALG} Transformer as the backbone model. We perform our training and evaluation based on the MMDetection\cite{chen2019mmdetection}. In the training process, we take AdamW optimizer with initial learning rate of 0.0001, weight decay of 0.05, and batch size of 2. Multi-scale  training\cite{carion2020end} strategy is also used to resize the input so that the shorter side is between 480 and 800 and the longer side is at most 1333. Following previous works, we evaluate the performance of our \emph{SALG} backbone model with two different training schedules: 1$\times$ schedule with 12 epochs and 3$\times$ schedule with 36 epochs. Different from classification, detection and segmentation models process input images with various sizes. We pad the image patches with zeros so that they can be partitioned by the given window size before unsupervised semantic-aware region segmentation.

\smallskip\noindent\textbf{Results on Object Detection and Instance Segmentation.}
We report the performance of different backbone models with Mask R-CNN~\cite{he2017mask} framework in Table~\ref{tab:table2}. Our \emph{SALG} Transformer shows superior performance over other Conv-based and Transformer-based backbone models with the advantages of proposed unsupervised semantic-aware region segmentation and local-global Transformer blocks. 
Applying local attention within multi-scale semantic-aware regions in different stages enables our \emph{SALG} Transformer to capture the fine-gained features of objects with different scales, which is crucial for dense-prediction tasks such as object detection and instance segmentation.

As shown in Table~\ref{tab:table2}, for both tasks of object detection and instance segmentation, our \emph{SALG} Transformer model as backbone outperforms Res50~\cite{he2016deep} and Res101~\cite{he2016deep} with comparable model size, demonstrating the effectiveness of our model over Conv-based models. For example, with 1$\times$ schedule, \emph{SALG}-T and \emph{SALG}-S  outperforms Res50~\cite{he2016deep} and Res101~\cite{he2016deep} by 5.7 $AP^b$ and 4.4 $AP^b$ respectively in the task of object detection. Besides, with 3$\times$ schedule, our model also achieves 5.5 $AP^b$ and 3.7 $AP^b$ over Res50 and Res101 respectively. Similar results can be observed in the experiments for instance segmentation.
Furthermore, compared with Swin Transformer~\cite{liu2021swin}, which shows strong capability in both object detection and instance segmentation tasks, our \emph{SALG} Transformer also shows its superior performance on smaller-scale variants including \emph{SALG}-ST and \emph{SALG}-T. 

\smallskip\noindent\textbf{Qualitative evaluation.}
We also perform qualitative comparison between our \emph{SALG} Transformer and Swin Transformer~\cite{liu2021swin}, in the `super-tiny' modeling scale, on dense-prediction tasks of object detection and instance segmentation by visualizing the class activation maps (CAM) on randomly selected samples from COCO dataset in Figure~\ref{fig:figure6}.

In COCO-2017 dataset, there are usually multiple objects to be detected in an image. To perform qualitative evaluation, we choose one specific class for Grad-CAM visualizations~\cite{selvaraju2017grad} in each image. 
We randomly choose four images from the validation set of COCO and visualize the class activation maps of the target classes. The results in Figure~\ref{fig:figure6} show that our model is able to capture the related objects in each image and thus predict the bounding box more precisely than Swin Transformer. By contrast, Swin Transformer makes several wrong prediction of bounding boxes (indicated by red bounding boxes) due to imprecise object detection (indicated by CAM) during feature learning. These results demonstrates the advantages of our model in dense-prediction tasks.


\subsection{Ablation Studies}
\label{sec:4.3}
In this section, we conduct a series of ablation studies for the task of image classification on ImageNet-1K dataset ~\cite{deng2009imagenet} to evaluate the effectiveness of our proposed components, including Semantic-Aware Region Segmentation and Local-Global Transformer block. Besides, we also compare the performance of two different ways of inter-region feature propagation discussed in Section~\ref{sec:3.3}, namely using Multi-head Self-Attention (MSA) and using averaging pooling (AP).

We perform ablation studies on four variants on our proposed \emph{SALG}-ST and \emph{SALG}-T models:

$\bullet$ \textbf{Base model}, which divides the feature map into regular windows and performs multi-head self-attention within these local windows. Hierarchical architecture is also employed to generate features with multiple scales. As a result, this variant is equivalent to Swin Transformer~\cite{liu2021swin} without shifted windowing scheme.

$\bullet$ \textbf{Base+Local\&Global}, which adopts the proposed Local-Global Transformer blocks on regular windows, to learn local features within each window and global dependencies between windows. Compared to our \emph{SALG}, no Semantic-Aware Region Segmentation is performed. Hence, this variant is proposed to validate the effectiveness of our devised Semantic-Aware Region Segmentation.

$\bullet$ \textbf{Base+Seg.+Local}, which performs Semantic-Aware Region Segmentation to obtain semantic regions, and adopts Intra-Region self-attention within each regions to learn fine-grained features. Compared to our \emph{SALG}, no Inter-Region Feature Propagation is performed. Hence, this variant is unable to learn global dependencies. 

$\bullet$ \textbf{SALG}, which is our intact model. Semantic-Aware Region Segmentation is applied to explore underlying semantic priors. Local-Global Transformer Block is leveraged to learn local object-wise semantics and global dependencies by Intra-Region Self-Attention and Inter-Region Feature Propagation, respectively. 

Table~\ref{tab:table3} presents the experimental results of the ablation studies on the two scales of our \emph{SALG}: `super tiny' and `tiny'. 

\smallskip\noindent\textbf{Effect of Local-Global Transformer Block.} For both `super tiny' and `tiny' versions, the performance is improved from \textbf{Base model} to \textbf{Base+Local\&Global}. It indicates the effectiveness of the proposed Local-Global Transformer Block, which learns fine-grained features and global dependencies alternately in an efficient manner. The performance gain from \textbf{Base+Seg.+Local} to our \textbf{SALG} also manifests the advantage of modeling global dependencies through inter-region feature propagation in the proposed Local-Global Transformer Block. 

\smallskip\noindent\textbf{Effect of Semantic-Aware Region Segmentation.} Our intact \textbf{SALG} outperforms \textbf{Base+Local\&Global} by $2\%$ and $0.4\%$ in terms of top-1 accuracy in `super-tiny' and `tiny' versions respectively. Meanwhile, the performance is significantly improved by $1.6\%$ and $2.1\%$ from \textbf{base model} to \textbf{Base+Seg.+Local}. It manifests the remarkable advantages of our proposed semantic-aware region segmentation, which enables the transformer model to perform self-attention on semantically meaningful regions instead of regular windows, to further learn more effective features. 

\smallskip\noindent\textbf{Comparison between `MSA' and `AP' for inter-region feature propagation.} As discussed in Section~\ref{sec:3.3}, we investigate two optional ways for inter-region feature propagation based on the AGR tokens, which aggregate representation of corresponding regions:
Multi-head Self-Attention (\textbf{MSA}) as in Equation \ref{eqn:agr-msa} and Averaging Pooling (\textbf{AP}) as in Equation \ref{eqn:agr-avg}. 

Table~\ref{tab:table4} presents the experimental results of these two ways on our \emph{SALG} with two scales: `super-tiny' and `tiny'.
For \emph{SALG}-ST, the method \textbf{MSA} outperforms \textbf{AP} by $0.7\%$ in terms of accuracy at the cost of increasing $0.6$ M model parameters. It indicates the method \textbf{MSA} can perform deeper interaction between AGR tokens, which is advantageous for the small-scale model with limited modeling capacity. 
For \emph{SALG-T} with higher modeling capacity, 
the performance gain from adopting the method of \textbf{MSA} for inter-region feature propagation is limited.

\begin{table}[!t]
\caption{Ablation studies of Semantic-Aware Region Segmentation and Local-Global Transformer Block.}
\label{tab:table3}
\centering
\resizebox{0.9\linewidth}{!}{
\begin{tabular}{c|l|c}
\toprule
\textbf{Model Scale} & \textbf{Variant} & \textbf{Top-1 ($\%$)} \\
\midrule
\multirow{4}{*}{Super-tiny} & Base model & 73.7\\
& Base + Local \& Global &  74.6\\
 & Base + Seg. + Local &  75.3 \\
 & SALG  &  76.6 \\
\midrule
\multirow{4}{*}{Tiny} & Base model & 80.2\\
 & Base + Local \& Global   & 82.0\\
 & Base + Seg. + Local  & 82.3\\
 & SALG    & 82.4\\
 
\bottomrule
\end{tabular}
}
\end{table}

\begin{table}[!t]
\caption{Comparison of two different implementations for inter-region feature propagation on different scales of \emph{SALG}.}
\label{tab:table4}
\centering
\resizebox{0.8\linewidth}{!}{
\begin{tabular}{c|c|c|c}
\toprule
\textbf{Model Scale} & \textbf{Method}  & \textbf{Params} & \textbf{Top-1($\%$)} \\
\midrule
\multirow{2}{*}{Super-tiny} & AP & 6.5M &  75.9\\
 & MSA & 7.1M & 76.6\\
\midrule
\multirow{2}{*}{Tiny} & AP & 32.2M &  82.1\\
& MSA & 42.4M & 82.4\\
\bottomrule
\end{tabular}}
\end{table}

%% file: Conclusion.tex
In this work, we have presented Semantic-Aware Local-Global (\emph{SALG}) Vision Transformer, which is the general-purpose backbone for vision transformer. Unlike typical vision transformers that perform uniform partition to produce equal size of windows for local self-attention, our \emph{SALG} performs semantic segmentation in an unsupervised way to explore underlying semantics priors in the image. Moreover, our \emph{SALG} learns features from the segmented semantic regions using the proposed local-global transformer bock, which performs both 1) local intra-region self-attention for learning fine-grained features within each segmented region, and 2) global inter-region feature propagation for modeling global dependencies among all regions. Consequently, our \emph{SALG} is able to learn features for each token efficiently in a global view, which is the essential advantage of Transformer. Extensive experiments across three tasks, including image classification, object detection and instance segmentation, demonstrate the advantages of our model over other state-of-the-art methods. In particular, our model is particularly advantageous in small-scale modeling scenarios when the modeling capacity is not sufficient for other models to learn semantics implicitly.